\documentclass{article}
\usepackage{spconf,amsmath,graphicx}
\usepackage{amssymb}
\usepackage{multirow}
\usepackage{color}

\usepackage[english]{babel}
\usepackage[utf8]{inputenc}
\usepackage{balance}
\usepackage{algorithm}
\usepackage[noend]{algpseudocode}

\title{Detection of Adversarial Attacks and Characterization of Adversarial Subspace}
\name{Mohammad Esmaeilpour, Patrick Cardinal, Alessandro Lameiras Koerich}
\address{\'{E}cole de Technologie Sup\'{e}rieure (\'{E}TS)\\
D\'{e}partement de G\'{e}nie Logiciel et des TI\\
1100 Notre-Dame W, Montr\'{e}al, H3C 1K3, Qu\'{e}bec, Canada}
\begin{document}
%
\maketitle
\begin{abstract}
Adversarial attacks have always been a serious threat for any data-driven model. 
In this paper, we explore subspaces of adversarial examples in unitary vector domain, and we propose a novel detector for defending our models trained for environmental sound classification. We measure chordal distance between legitimate and malicious representation of sounds in unitary space of generalized Schur decomposition and show that their manifolds lie far from each other. Our front-end detector is a regularized logistic regression which discriminates eigenvalues of legitimate and adversarial spectrograms. The experimental results on three benchmarking datasets of environmental sounds represented by spectrograms reveal high detection rate of the proposed detector for eight types of adversarial attacks and outperforms other detection approaches.
\end{abstract}


\section{Introduction}
\label{sec:intro}
In the field of sound and speech processing, it is very common to use 2D representations of audio signals for training data-driven algorithms for both regression and classification tasks. Such 2D representations have lower dimensionality than audio waveforms and they easily fit advanced deep learning architectures mainly developed for computer vision applications. Mel frequency cepstral coefficient (MFCC), short-time Fourier transformation (STFT), discrete wavelet transformation (DWT) are among the most pervasive 2D signal representations which essentially visualize frequency-magnitude distribution of a given reconstructed signal over time. Thus far, the best sound classification accuracy has been achieved for deep learning algorithms trained on 2D signal representations \cite{boddapati2017classifying,esmaeilpour2019unsupervised}. However, it has been shown that despite achieving high performance, the approaches based on 2D representations are very vulnerable against adversarial attacks \cite{esmaeilpour2019robust}. Unfortunately, this poses a strict security issue because crafted adversarial examples not only mislead the target model toward a wrong label, but also, they are transferable to other models including conventional algorithms such as support vector machines (SVM) \cite{esmaeilpour2019robust}.

There are some discussions about existence, origin, and behavior of adversarial examples, notably their linear characteristics \cite{goodfellow2014explaining}, but there is no reliable approach to discriminate their underlying subspace(s) compared to legitimate examples. In an effort to characterize possible adversarial subspaces, some detectors have been introduced. They are mainly based on a statistical comparison on predictions of the victim model. Feinman et al.~\cite{feinman2017detecting} have proposed to estimate kernel density (KD) and Bayesian uncertainty (BU) of a trained deep neural network (DNN) for triplets of legitimate, noisy, and adversarial examples. All these measurements have been carried out with the assumption of approximating a DNN to a deep Gaussian process and they result in high ratios of KD and BU for adversarial examples compared to legitimate and noisy samples. Measuring maximum mean discrepancy and energy distance of examples are two other statistical metrics for investigating adversarial manifolds using divergence of model predictions for clusters of datapoints \cite{grosse2017statistical}. In addition to these output-level statistical measurements, logits of adversariality have been carefully assessed in each subnetwork placed on top of some hidden units of the victim model \cite{metzen2017detecting} as well as measuring instability of potential layers to perturbations~\cite{rouhani2017curtail}. Ma et al.~\cite{ma2018characterizing} presented a comprehensive study for characterizing adversarial manifolds and introduced local intrinsic dimensionality (LID) score, which measures $\ell_{2}$ distance of network prediction for a given example compared to prediction logits of its $k$ neighbours at each hidden unit. The actual detector is a logistic regression binary classifier trained on one class made up of LID vectors of legitimate and noisy examples because they lie in a very close subspace and another class made up of LID vectors of adversarial examples generated by strong attacks. Experimental results on several datasets have shown the competitive performance of the LID detector compared to KD and BU \cite{ma2018characterizing}. Unfortunately, it has been shown that these detectors of adversarial examples might fail to detect strong adversarial attacks in adverse scenarios \cite{carlini2017adversarial, athalye2018obfuscated}, due to the difficulty in tuning detectors or even due to the particular characteristics of the datasets.  

In this paper we show that, adversarial manifolds lie far from legitimate and noisy examples using a unitary space-based chordal distance metric. We also provide an algorithm for proactively detect potential malicious examples using generalized Schur decomposition (a.k.a. QZ decomposition) \cite{van1983matrix}. This paper is organized as follows. Section~\ref{sec:adv_detect} presents a brief explanation of unitary space of QZ as well as our adversarial detection algorithm. Experimental results on DWT representation of three environmental sound datasets are discussed in the Section~\ref{sec:exp}. Conclusions and perspectives of future work are presented in the last section.

\section{Adversarial Detection}
\label{sec:adv_detect}
Computing norm metrics is a common approach for measuring the similarity between crafted adversarial examples and their legitimate counterparts. In addition to basic norms such as $l_{2}$ and $l_{\infty}$, human visual inference oriented metric has been also embedded in general optimization problems \cite{rozsa2016adversarial}. These similarity constraints are probably the most valuable clues in studying possible subspaces of crafted examples.

It has been shown that, regardless of the category or type of adversarial attack, the generated examples, subject to a similarity constraint, lie in a sub-Cartesian space further than the legitimate ones \cite{ma2018characterizing}. However, this is tricky and may not work correctly for strong attacks \cite{athalye2018obfuscated}. Our detailed study of the failure cases of such detectors uncovered imperfection of Cartesian metric space (distance-based) for exploring adversarial subspaces. Therefore, vector spaces that may discriminate between adversarial and legitimate manifolds can be very useful to build robust adversarial example detectors.

We investigate in this paper the mapping of input samples to the vector space of generalized Schur decomposition and the use of chordal distance to identify their underlying subspaces.

\subsection{Schur Decomposition and Chordal Distance}
\label{sec:chordal_schur}
For computing generalized Schur decomposition of two spectrograms denoted as $M_{1}$ and $M_{2}$ in a complex set $\mathbb{C}^{n \times n}$ there should exist unitary matrices $Q$ and $Z$ such that:
\begin{equation}
    Q^{H}M_{1}Z = T, \quad Q^{H}M_{2}Z = S
\end{equation}
\noindent where $S$ and $T$ are upper (quasi) triangular and $Q^H$ denotes the conjugate transpose of $Q$. The eigenvalues ($\lambda$) of $M_{1}$ and $M_{2}$ can be approximated as:
\begin{equation}
    \lambda (M_{1}, M_{2}) = \{ t_{ii}/s_{ii} : s_{ii} \neq 0 \}
    \label{eigen_lambda}
\end{equation}

\noindent where $t_{ii}$ and $s_{ii}$ are diagonal elements of $T$ and $S$, respectively, and $\lambda (M_{1}, M_{2}) = \mathbb{C}$ for some zero-valued diagonal entries of $S$ and $T$. In other words, super-resolution similarity between two spectrograms can be calculated as: 
\begin{equation}
    \mathrm{det}(M_{1}-\lambda M_{2})=\mathrm{det}
\begin{pmatrix}
QZ^{H}
\end{pmatrix}
\prod_{i=1}^{n}(t_{ii}-\lambda s_{ii})
    \label{detmm}
\end{equation}

\noindent Implied by Bolzano-Weierstrass theorem \cite{van1983matrix}, the bounded basis matrices of $\begin{Bmatrix} (Q_{k}, Z_{k}) \end{Bmatrix}$ support $\lim_{i\rightarrow \infty} (Q_{k}, Z_{k})=(Q,Z)$. The unitary subsequence of $Z_{k}$ leads to the following relation:
\begin{equation}
    Z_{k}^{H}(M_{2,k}^{-1}Q_{k})=S_{k}^{-1}
    \label{det_sim}
\end{equation}

\noindent which asymptotically implies $Q_{k}^{H}\begin{pmatrix} M_{1}M_{2}^{-1}Q_{k} \end{pmatrix}$ equivalent to generic Schur decomposition of $M_{1}M_{2,k}^{-1}$ for nonsingular basis matrices of $\begin{Bmatrix} M_{2,k} \end{Bmatrix}$.

In perturbing the spectrogram $M_{i}$ where $\widetilde{M_{i}}\simeq  M_{i}+\epsilon$  increases considerably the chance of noticeable variations in the resulting eigenvalues/eigenvectors \cite{van1983matrix}. Theoretically, we can measure it using the chordal metric where the pencil of $\vec{\mu}_{i} M_{i}-\widetilde{M_{i}}$ is the point of interest for $\mu_{i} \in \begin{Bmatrix} \begin{pmatrix} t_{ii}, s_{ii} \end{pmatrix}| s_{ii}/t_{ii} \end{Bmatrix}$ perturbed by $\epsilon$ as conditioned in Eq.~\ref{condition_qz}.
\begin{equation}
\left \| M_{i}-\widetilde{M_{i}} \right \|_{2} \simeq \epsilon_{i}
\label{condition_qz}
\end{equation}

\noindent where $\epsilon_{i}$ is a very small perturbation. The chordal distance for the vectors of eigenvalues associated with pencil of $\vec{\mu}_{i} M_{i}-\widetilde{M_{i}}$ can be measured by Eq.~\ref{mychord_dist} \cite{van1983matrix}. 
\begin{equation}
    \mathrm{chord}(\lambda_{i}, \lambda_{i,\epsilon})=\frac{\left | \lambda_{i} - \lambda_{i,\epsilon} \right |}{\sqrt{1+\lambda_{i}^{2}}\sqrt{1+\lambda_{i,\epsilon}^{2}}} 
    \label{mychord_dist}
\end{equation}

\noindent where pencils are neither necessarily bound to be normalized nor differentiable. For any adversarial attack that perturbs a legitimate spectrogram $M_{i}$ by $\epsilon_{i}$, we compute chordal distance as Eq.~\ref{mychord_dist} and we compare the distances obtained to find separable manifolds for legitimate and adversarial examples.

\subsection{Adversarial Subspace}
\label{adv_subspace}
We can explore the properties of adversarial examples using chordal distance in unitary space of eigenvectors where each spectrogram is represented by basis functions $Q_{i}$ and $Z_{i}$. For any legitimate and adversarial spectrograms, the chordal distance between their associated eigenvalues ($\lambda , \lambda_{i,\epsilon}$) must satisfy the constraint defined in Eq.~\ref{chord} \cite{van1983matrix}.
\begin{equation}
    \mathrm{chord}(\lambda_{i}, \lambda_{i,\epsilon})\leq \frac{\epsilon}{\sqrt{\begin{bmatrix}
\begin{pmatrix}y^{H}M_{i}x \end{pmatrix}+\begin{pmatrix} y^{H}\widetilde{M_{i}}x \end{pmatrix}\end{bmatrix}^2}}
\label{chord}
\end{equation}

\noindent where $x$ and $y$ satisfy $M_{i}x=\lambda \widetilde{M_{i}}x$ and $y^{H} M_{i} = \lambda y^{H}\widetilde{M_{i}}$ for the symmetric in the upper bound of $M_{i}$ and $\widetilde{M_{i}}$. The extreme case for the defined pencil may happen when both $s_{ii}$ and $t_{ii}$ are zero. Therefore, we can replace their division with a small random value close to their neighbours.

Not only satisfying Eq.~\ref{condition_qz} is required for properly computing chordal distance of eigenvalues, but it must also be part of the optimization procedure of any adversarial attack because the perturbation value $\epsilon$ should not be perceivable. For adversarial perturbations, an adjustment of the chordal distance by a factor $\gamma$ is also required ($\mathrm{chord}(\lambda_{i}, \lambda_{i,\epsilon}) + \gamma_{i}$). The value of such a hyperparameter should be very small and associated to mean eigenvalue, otherwise it might cause ill-conditioning cases. We examine the effects of different pencil perturbations on the chordal distance and inequality of Eq.~\ref{chord} from random noisy to carefully optimized adversarials in Section~\ref{sec:exp}.

\subsection{Adversarial Discrimination}
In practice, detecting adversarial examples using chordal distance for a given test input requires access to its reference spectrogram as well as to the perturbation $\epsilon$. However, this is not feasible for real life applications. For rectifying this issue, we propose to compare eigenvalues of legitimate and adversarial examples to draw a decision boundary between them. To this end, we train a logistic regression on the eigenvalues of legitimate and adversarial examples as shown in Algorithm~\ref{alg:QZdecomposition}.  

For every spectrogram pairs randomly picked from an identical class, we compute their associated eigenvalues using QZ decomposition. We assume that spectrograms have been generated for short audio signals and they share significant similarities, especially when they are split into smaller batches.
\begin{algorithm}[ht]
     \caption{Discriminating adversarial examples from legitimate ones using their associated eigenvectors.
     }
     \label{alg:QZdecomposition}
     \label{pseudo}
     \begin{algorithmic}[1] \label{qzadv_pcode}
     \small
      \Require {$C_{leg}$: Class of legitimate samples}
       \Ensure{Detector$\begin{bmatrix} \mathrm{schur(M)} \end{bmatrix}$} \Comment{$M$: test spectrogram}
       \State $\Lambda_{leg} = []$ , $\Lambda_{adv}=[]$ \Comment{lists}
       \For{$B_{leg}$ in $C_{leg}$} \Comment{$B_{leg}$: legitimate batch}
       \State $B_{adv} :=$ adversarial attack on $B_{leg}$\newline \hfill\rlap{\hspace*{1em}}{\Comment{$B_{adv}$: adversarial batch}}
       \State $\overrightarrow{\lambda}_{leg} = \mathrm{eigen} \begin{bmatrix}
        \mathrm{qz}\begin{pmatrix}B_{leg}\begin{bmatrix} i \end{bmatrix},B_{leg}\begin{bmatrix} j \end{bmatrix}\end{pmatrix}
        \end{bmatrix}$ \Comment{$i \neq j$}
        
       \State $\overrightarrow{\lambda}_{adv} = \mathrm{eigen} \begin{bmatrix}
        \mathrm{qz}\begin{pmatrix}B_{adv}\begin{bmatrix} i \end{bmatrix},B_{adv}\begin{bmatrix} j \end{bmatrix}\end{pmatrix}
        \end{bmatrix}$ \Comment{$i \neq j$}

       \State $\Lambda_{leg}.\mathrm{append}\begin{pmatrix}\overrightarrow{\lambda}_{leg}\end{pmatrix}$, $\Lambda_{adv}.\mathrm{append}\begin{pmatrix}\overrightarrow{\lambda}_{adv}\end{pmatrix}$
       \EndFor
       \State Detector$\begin{bmatrix} \mathrm{schur(M)} \end{bmatrix}$ = train a classifier on ($\Lambda_{leg}$, $\Lambda_{adv}$)

     \end{algorithmic}
\end{algorithm}

For a test input spectrogram $M$, its eigenvalues generated by Schur decomposition will be used as arguments for the final front-end classifier (the detector) as relations of these two decomposition have been explained in Section \ref{sec:chordal_schur}. Generalizing this algorithm to a multiclass classification problem requires computing eigenvectors of inter-class samples sharing no significant similarity due to causing ill-conditioned decomposition for pencils. 
\section{Experimental Results}
\label{sec:exp}
\begin{table*}[htp]
\centering
\small
\caption{The mean $\gamma$ values for justifying chordal distances of adversarial examples, the corresponding mean perturbation and the recognition accuracy of victim models (ConvNet \& SVM) on adversarial sets.}
\begin{tabular}{|c|c|c|c|c|c|c|c|c|}
\hline
              & FGSM         & BIM-a        & BIM-b         & JSMA         & CWA           & Opt          & EA           & LFA           \\ \hline
$\gamma$      & $7 \pm 0.12$ & $6 \pm 0.03$ & $8 \pm 0.017$ & $7 \pm 0.17$ & $11 \pm 0.09$ & $10 \pm 0.27$ & $8 \pm 0.39$ & $12 \pm 0.16$ \\ \hline
$\ell_{2}$       & $5.637$      & $4.015$      & $6.371$       & $6.187$      & $4.426$       & $5.067$      & NA           & NA            \\ \hline
Accuracy (\%) & $3.036$      & $6.017$      & $4.964$       & $3.189$      & $6.237$       & $8.143$       & $15.157$      & $17.845$       \\ \hline
\multicolumn{9}{l}{\scriptsize NA: Not Applicable.}
\end{tabular}
\label{gamma_adjust}
\end{table*}

\begin{table*}[htp]
\centering
\small
\caption{Mean class-wise comparison of the AUC (\%) achieved by the adversarial detectors for spectrograms attacked with eight adversarial attacks. The best results are highlighted in bold.}
\begin{tabular}{|c|c|c|c|c|c|c|c|c|}
\hline
& \multicolumn{6}{|c|}{ConvNet} & \multicolumn{2}{|c|}{SVM} \\ \cline{2-9}
Detector & FGSM & BIM-a & BIM-b & JSMA & CWA & Opt & EA & LFA \\ \hline
KD       &   $ 65.234 $   &   $ 88.097 $    &  $  87.914$     &   $ 63.552 $   &  $ 61.025 $   &    $  86.105$  &   $ 55.479 $ &  $ 63.659 $   \\ \hline
BU       &  $ 39.025 $    &  $ 80.673 $     &   $ 55.474 $    &     $  80.603$ &  $ 58.022 $   &   $ 69.207 $   &  $  57.861$  &   $  67.610$  \\ \hline
KD+BU    &  $ 74.381 $    &  $ 91.154 $     &   $  88.243$    &  $ 89.251 $    &   $ 64.349 $  &  $  90.461$    & $ 58.330 $   &   $ 69.008 $  \\ \hline
LID      &  $ 79.299 $    & $ 93.097 $      &   $ 94.671 $    & $  91.665$     &   $ 75.297  $  &  $  \textbf{94.781}$    & $ 70.981 $   &  $ \textbf{71.239} $   \\ \hline
Proposed &  $ \textbf{84.132} $    & $ \textbf{96.519} $      &   $ \textbf{95.349} $    &     $ \textbf{94.375} $ &    $ \textbf{89.957} $ & $  93.309$     & $ \textbf{75.227} $   &  $ 71.198 $   \\ \hline
\end{tabular}
\label{final_res}
\end{table*}

In this section, we study the performance of computing chordal distance on adversarial detection and we evaluate the performance of the proposed detector in adverse scenarios on three environmental sound datasets: ESC-10 \cite{DVN/YDEPUT_2015}, ESC-50 \cite{DVN/YDEPUT_2015}, and UrbanSound8k \cite{Salamon:UrbanSound:ACMMM:14}. The first dataset includes 400 five-second length audio recordings of 10 classes. It is actually a simplified version of ESC-50 which has 2000 samples of 50 classes with the same length. The UrbanSound8k dataset contains 8732 samples ($\leq 4$s) of 10 classes and compared to the first two datasets, it provides more sample diversity both in terms of quality and quantity.

We apply pitch-shifting operation as part of 1D signal augmentation as proposed in \cite{esmaeilpour2019unsupervised}. This low-level data augmentation increases the chance of learning more discriminant features by the classifier, especially for ESC-10 and ESC-50 compared to UrbanSound8k. Four pitch-shifting scales, namely $0.75, 0.9, 1.15, 1.5$ are applied to each sample in order to add four new samples to the legitimate sets. These hyperparameters are reported to be the most effective scales for the benchmarking datasets \cite{esmaeilpour2019unsupervised}. The wavelet mother function which we use for producing DWT spectrogram representations is complex Morlet. Sampling frequencies and frame length are set to 8 kHz and 50 ms for ESC-10 and UrbanSound8k and 16 kHz and 30 ms for ESC-50 with fixed overlapping ratio of 0.5 for all datasets~\cite{boddapati2017classifying}. The convolution of the Morlet function with the signal produces a complex function with considerable overlap between real and imaginary parts. Therefore, for representing real spectrograms we use linear, logarithmic, and logarithmic real visualizations. The first visualization scheme highlights high-frequency magnitudes which denote high variation areas. Low-frequency information has been characterized by a logarithmic operation which expands their distances. Energy of the signal, which is associated with the signal’s mean, has been obtained by applying a logarithmic filter on the real part.

Since the frequency-magnitude of a signal distributed over time has variational dimensions, none of the three mentioned visualizations produce square spectrograms. Hence, we bilinearly interpolate each spectrogram to fit square size with respect to this constraint of QZ decomposition. The actual size of the spectrograms for ESC-10 and ESC-50 is 1536$\times$768 and 1168$\times$864 for UrbanSound8k because the latter has shorter audio recordings of at least one second. Final size of spectrogram after downsampling and interpolation is 768$\times$768. This lossy operation may remove some pivotal frequency information and consequently it may decrease the performance of the classifier. However, obtaining the highest recognition accuracy is not our point of interest in this paper, but studying adversarial subspaces.

For the choice of the victim classifier, we use an SVM and a convolutional neural network (ConvNet) to compare detection rate of the proposed detector for variety of adversarial attacks. In SVM configuration, we use scikit-learn \cite{scikit-learn} with a grid search. Linear, polynomial, and RBF kernels have been tested on the 2/3 of the shuffled datasets (training and development). The best recognition accuracy on the test set was achieved with the RBF kernel with about 72.056\%, 71.257\%, 72.362\% for ESC-10, ESC-50 and UrbanSound8k datasets, respectively. The proposed ConvNet has four convolutional layers with receptive field 3$\times$3, stride 1$\times$1, and 128, 256, 512, and 128 filters, respectively. On top of the last convolution layer there are two fully connected layers of sizes 256 and 128. All layers use ReLU activation function, except the output layer for which softmax is used. Batch and weight normalization have been applied at all convolutional layers. Such a ConvNet can achieve recognition performance of 73.415\%, 73.674\%, and 75.376\% for ESC-10, ESC-50, and UrbanSound8k datasets respectively on the 1/3 test set.

We attack the ConvNet by fast gradient sign method (FGSM)~\cite{goodfellow2014explaining}, basic iterative methods (BIM-a and BIM-b)~\cite{kurakin2016adversarial}, Jacobian-based salience map attack (JSMA)~\cite{papernot2016limitations}, optimization-based attack (Opt)~\cite{liu2016delving}, and Carlini \& Wagner attack (CWA)~\cite{carlini2017towards}. For the SVM model, we use label flipping attack (LFA)~\cite{xiao2012adversarial} and evasion attack (EA)~\cite{biggio2013evasion}. Overall, for each legitimate DWT spectrogram ($M_{i}$), eight adversarial examples are crafted ($\widetilde{M_{i,j}}$ for $j=1\dots 8$). For each pencil of $\mu_{i} M_{i}-\widetilde{M_{i,j}}$, we measure their chordal distances using Eq.~\ref{mychord_dist}, then for a random unit 2-norm $x$ and $y$ matrices, we check for the inequality as stated in Eq.~\ref{chord} and required $\gamma$ adjustments. Similarly, we add random Gaussian noise to each $M_{i}$ with zero mean and  $\sigma\in\begin{Bmatrix} 0.01, 0.02, 0.04, 0.05 \end{Bmatrix}$ and build pencil of $\mu_{i} M_{i}-N_{i,k}$ where $N_{i,j}$ for $k=1\dots 4$ denote the noisy spectrograms which also satisfy Eq.~\ref{condition_qz}. Table~\ref{gamma_adjust} summarizes the adjustment of $\gamma$ required for crafted adversarial examples to satisfy Eq.~\ref{chord}. For the generated noisy samples, an adjustment of $\gamma$ to $0.5 \pm 0.012$ is needed, which is averaged over different values of $\sigma$. Considerable displacement between chordal distance adjustments required for adversarial and noisy spectrogram sets denote their non-identical and dissimilar subspaces.   

For testing the performance of the Algorithm~\ref{alg:QZdecomposition} in discriminating adversarial from legitimate examples, we use all the attacks mentioned above for crafting $B_{adv}$. Regularized logistic regression has been used as the front-end classifier for discriminating $\Lambda_{leg}$ from $\Lambda_{adv}$. We compare the performance of the proposed detector versus LID, KD, BU, and the combination KD+BU. Table~\ref{final_res} shows that the proposed detector outperforms other detectors for the majority of the attacks. The proposed detector can be used for MFCC and STFT representations of sounds or even other datasets commonly used for computer vision applications. The key challenge in this detector is its sensitivity to intra-class sample similarities, otherwise it may not satisfy Eq.\ref{chord}, especially for black-box multiclass discrimination.

\section{Conclusion}
Since adversarial examples are visually very similar to the legitimate samples, differentiating their underlying subspaces is very challenging in metric space of Cartesian. In this paper we showed that offset between subspace of legitimate spectrograms compared to their associated adversarial examples can be measured by a metric called chordal distance defined in unitary vector space of generalized Schur decomposition. Using this metric, we demonstrated that manifold of adversarial examples lie far from legitimates and noisy samples which have been slightly perturbed by Gaussian filters. 

In order to detect any adversarial attacks when there is no access neither to reference spectrogram nor adversarial perturbation, we proposed a detector which is a regularized logistic regression model for discriminating eigenvalues of a malicious spectrogram from legitimates. Experimental results on three benchmarking environmental sound datasets showed our proposed detector outperforms other detectors for six out of eight different adversarial attacks.
For future studies, we would like to improve chordal distance to better characterize adversarial manifolds and also study possibility of encoding this metric directly into the adversarial detector.

\vfill\pagebreak
\balance
\bibliographystyle{IEEEbib}

\begin{thebibliography}{10}

\bibitem{boddapati2017classifying}
Venkatesh Boddapati, Andrej Petef, Jim Rasmusson, and Lars Lundberg,
\newblock ``Classifying environmental sounds using image recognition
  networks,''
\newblock {\em Procedia Computer Science}, vol. 112, pp. 2048--2056, 2017.

\bibitem{esmaeilpour2019unsupervised}
M.~Esmaeilpour, P.~Cardinal, and A.~L. Koerich,
\newblock ``Unsupervised feature learning for environmental sound
  classification using cycle consistent generative adversarial network,''
\newblock {\em arXiv preprint arXiv:1904.04221}, 2019.

\bibitem{esmaeilpour2019robust}
M.~Esmaeilpour, P.~Cardinal, and A.~L. Koerich,
\newblock ``A robust approach for securing audio classification against
  adversarial attacks,''
\newblock {\em arXiv preprint arXiv:1904.10990}, 2019.

\bibitem{goodfellow2014explaining}
Ian~J Goodfellow, Jonathon Shlens, and Christian Szegedy,
\newblock ``Explaining and harnessing adversarial examples,''
\newblock {\em arXiv preprint arXiv:1412.6572}, 2014.

\bibitem{feinman2017detecting}
Reuben Feinman, Ryan~R Curtin, Saurabh Shintre, and Andrew~B Gardner,
\newblock ``Detecting adversarial samples from artifacts,''
\newblock {\em arXiv preprint arXiv:1703.00410}, 2017.

\bibitem{grosse2017statistical}
Kathrin Grosse, Praveen Manoharan, Nicolas Papernot, Michael Backes, and
  Patrick McDaniel,
\newblock ``On the (statistical) detection of adversarial examples,''
\newblock {\em arXiv preprint arXiv:1702.06280}, 2017.

\bibitem{metzen2017detecting}
Jan~Hendrik Metzen, Tim Genewein, Volker Fischer, and Bastian Bischoff,
\newblock ``On detecting adversarial perturbations,''
\newblock {\em arXiv preprint arXiv:1702.04267}, 2017.

\bibitem{rouhani2017curtail}
Bita~Darvish Rouhani, Mohammad Samragh, Tara Javidi, and Farinaz Koushanfar,
\newblock ``Curtail: Characterizing and thwarting adversarial deep learning,''
\newblock {\em arXiv preprint arXiv:1709.02538}, 2017.

\bibitem{ma2018characterizing}
Xingjun Ma, Bo~Li, Yisen Wang, Sarah~M Erfani, Sudanthi Wijewickrema, Michael~E
  Houle, Grant Schoenebeck, Dawn Song, and James Bailey,
\newblock ``Characterizing adversarial subspaces using local intrinsic
  dimensionality,''
\newblock {\em arXiv preprint arXiv:1801.02613}, 2018.

\bibitem{carlini2017adversarial}
Nicholas Carlini and David Wagner,
\newblock ``Adversarial examples are not easily detected: Bypassing ten
  detection methods,''
\newblock in {\em Proceedings of the 10th ACM Workshop on Artificial
  Intelligence and Security}. ACM, 2017, pp. 3--14.

\bibitem{athalye2018obfuscated}
A.~Athalye, N.~Carlini, and D.~Wagner,
\newblock ``Obfuscated gradients give a false sense of security: Circumventing
  defenses to adversarial examples,''
\newblock {\em arXiv preprint arXiv:1802.00420}, 2018.

\bibitem{van1983matrix}
Charles~F Van~Loan and Gene~H Golub,
\newblock {\em Matrix computations},
\newblock Johns Hopkins University Press, 1983.

\bibitem{rozsa2016adversarial}
A.~Rozsa, E.~M. Rudd, and T.~E. Boult,
\newblock ``Adversarial diversity and hard positive generation,''
\newblock in {\em Proc. IEEE Conf on Computer Vision and Pattern Recognition
  Workshops}, 2016, pp. 25--32.

\bibitem{DVN/YDEPUT_2015}
Karol~J. Piczak,
\newblock ``{ESC: Dataset for Environmental Sound Classification},'' 2015.

\bibitem{Salamon:UrbanSound:ACMMM:14}
J.~Salamon, C.~Jacoby, and J.~P. Bello,
\newblock ``A dataset and taxonomy for urban sound research,''
\newblock in {\em 22nd {ACM} Intl Conf on Multimedia}, Orlando, FL, USA, Nov.
  2014.

\bibitem{scikit-learn}
F.~Pedregosa, G.~Varoquaux, A.~Gramfort, V.~Michel, B.~Thirion, O.~Grisel,
  M.~Blondel, P.~Prettenhofer, R.~Weiss, V.~Dubourg, J.~Vanderplas, A.~Passos,
  D.~Cournapeau, M.~Brucher, M.~Perrot, and E.~Duchesnay,
\newblock ``{Scikit-learn: Machine Learning in Python },''
\newblock {\em Journal of Machine Learning Research}, vol. 12, pp. 2825--2830,
  2011.

\bibitem{kurakin2016adversarial}
Alexey Kurakin, Ian Goodfellow, and Samy Bengio,
\newblock ``Adversarial examples in the physical world,''
\newblock {\em arXiv preprint arXiv:1607.02533}, 2016.

\bibitem{papernot2016limitations}
Nicolas Papernot, Patrick McDaniel, Somesh Jha, Matt Fredrikson, Z~Berkay
  Celik, and Ananthram Swami,
\newblock ``The limitations of deep learning in adversarial settings,''
\newblock in {\em 2016 IEEE European Symposium on Security and Privacy
  (EuroS\&P)}. IEEE, 2016, pp. 372--387.

\bibitem{liu2016delving}
Yanpei Liu, Xinyun Chen, Chang Liu, and Dawn Song,
\newblock ``Delving into transferable adversarial examples and black-box
  attacks,''
\newblock {\em arXiv preprint arXiv:1611.02770}, 2016.

\bibitem{carlini2017towards}
Nicholas Carlini and David Wagner,
\newblock ``Towards evaluating the robustness of neural networks,''
\newblock in {\em IEEE Symp Secur Priv}, 2017, pp. 39--57.

\bibitem{xiao2012adversarial}
Han Xiao, Huang Xiao, and Claudia Eckert,
\newblock ``Adversarial label flips attack on support vector machines.,''
\newblock in {\em ECAI}, 2012, pp. 870--875.

\bibitem{biggio2013evasion}
Battista Biggio, Igino Corona, Davide Maiorca, Blaine Nelson, Nedim
  {\v{S}}rndi{\'c}, Pavel Laskov, Giorgio Giacinto, and Fabio Roli,
\newblock ``Evasion attacks against machine learning at test time,''
\newblock in {\em Joint European conference on machine learning and knowledge
  discovery in databases}. Springer, 2013, pp. 387--402.

\end{thebibliography}

\end{document}